\documentclass[review]{elsarticle}
\usepackage{lineno,hyperref}
\usepackage{amsmath}
\usepackage{graphicx}
\usepackage{subfigure}
\usepackage{caption}
\usepackage{amsfonts}
\usepackage{bm}
\usepackage{multirow}
\usepackage{pifont}%
\usepackage{algorithm}
\usepackage{algpseudocode}
\usepackage{color}
\usepackage{xcolor}

\def\eg{\textit{e.g.}}
\def\ie{\textit{i.e.}}

\def\etal{\textit{et al.}}

\modulolinenumbers[5]

\begin{document}

\begin{frontmatter}

\title{Low-resolution Human Pose Estimation}

\author[wang]{Chen Wang}
\ead{wangchen100@163.com}

\author[zhang]{Feng Zhang}
\ead{zhangfeng.ac@outlook.com}

\author[zhu]{Xiatian Zhu} 
\ead{eddy.zhuxt@gmail.com}

\author[ge]{Shuzhi Sam Ge}
\ead{samge@uestc.edu.cn}

\address[wang]{School of Computer Science and Engineering, University of Electronic Science and Technology of China, Chengdu, China.}
\address[zhang]{School of Computer Science and Technology, Nanjing University of Posts and Telecommunications, Nanjing, China.}
\address[zhu]{
Centre for Vision, Speech and Signal Processing,
Faculty of Engineering and Physical Sciences, University of Surrey, Guildford GU2 7XH, United Kingdom.}
\address[ge]{Department of Electrical and Computer Engineering, National University of Singapore, Singapore.}

\begin{abstract}
Human pose estimation has achieved significant progress 
on images with high imaging resolution. 
However, low-resolution imagery data bring nontrivial
challenges which are still under-studied.
To fill this gap,
we start with investigating existing methods and reveal that
the most dominant heatmap-based methods would suffer more 
severe model performance degradation from low-resolution,
and offset learning is an effective strategy.  
Established on this observation, in this work we propose a novel {\em Confidence-Aware Learning} (CAL) method which further addresses two fundamental limitations of existing 
offset learning methods: inconsistent training and testing,
decoupled heatmap and offset learning.
Specifically, 
CAL selectively weighs the learning of 
heatmap and offset with respect to ground-truth and 
most confident prediction, whilst capturing the statistical 
importance of model output in mini-batch learning manner.
Extensive experiments conducted on the COCO benchmark
show that our method outperforms significantly the state-of-the-art methods for low-resolution human pose estimation. 
\end{abstract}


\end{frontmatter}

\section{Introduction}

As a fundamental computer vision problem, human pose estimation is a critical capability in many tasks such as action recognition  \cite{CheronLS15}, human body generation \cite{0054HNBT20}, person re-identification \cite{QianFXWQWJX18}, pedestrian tracking \cite{AndrilukaRS10}, human-computer interaction \cite{jain2011real}, 
object detection \cite{centernet:zhou2019}, etc.
Human pose estimation is challenging due to large scale
variation, body occlusion, various shooting angles, and crowd scenes. 
A large number of approaches have been proposed \cite{hourglass:NewellYD16,CPN:ChenWPZYS18,cpm:WeiRKS16,attention:ChuYOMYW17,simplebaseline:XiaoWW18,HRNet:0009XLW19,UDP:huang2019devil,GRMI:PapandreouZKTTB17}. %
Typically, existing methods assume {\em high image resolutions},
\eg, more than 256$\times$256 in pixel,
whilst low-resolution (\eg{} 128$\times$96) images are largely neglected.
Although high-resolution input can provide more information for the human pose estimation task if available, person images can be captured only at low resolutions in many practical applications, for instance, wide-view video surveillance or long-distance shooting. 
In addition, high-resolution input would bring great computational and memory complexity, which hinders the pace of practical applications.
Therefore, {\em low-resolution human pose estimation} is a critical yet more challenging.

An intuitive approach for low-resolution human pose estimation is to recover the image resolution, \eg, by applying super-resolution methods \cite{sr:ledig2017photo,sr:wang2018esrgan,sr:zhang2018image,sr:soh2019natural,spsr:ma2020structure} as image pre-processing.
However, we find that the optimization of super-resolution
is not dedicated for high-level human pose analysis. In particular, the structural distortions and artifacts in super-resolved images often produced by existing super-resolution techniques are likely to deteriorate the performance of human pose models. Further, applying extra super-resolution increases both the computational overhead and pipeline complexity.
\begin{figure}[t]
    \centering
    \includegraphics[width=3.5in]{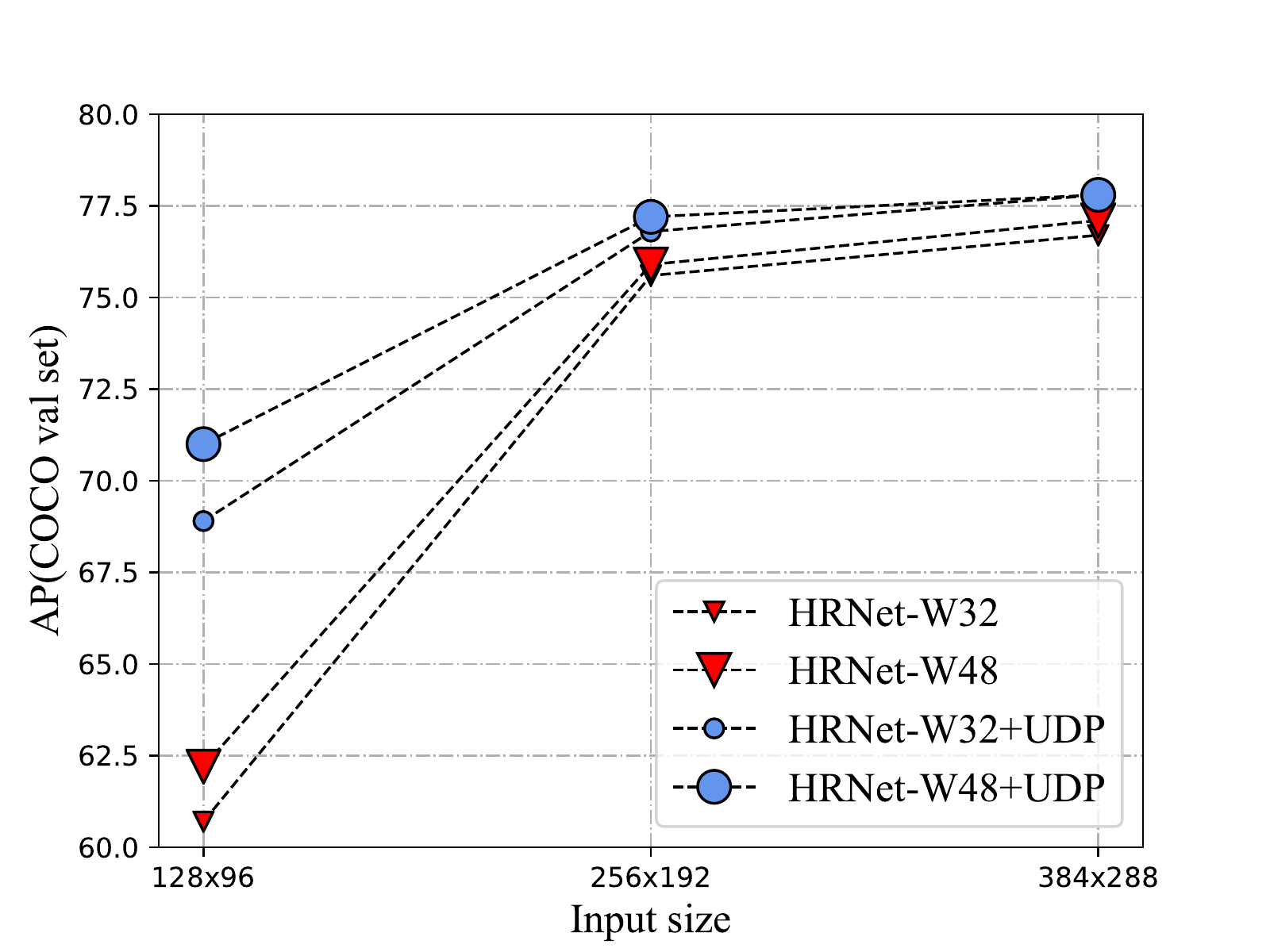}
    \caption{The performance of heatmap based HRNet \cite{HRNet:0009XLW19} and offset-based UDP \cite{UDP:huang2019devil} with different input sizes on COCO val set.}
    \label{fig:low_resolution}
\end{figure}

\begin{figure}[h]
    \centering
    \includegraphics[width=4.7in]{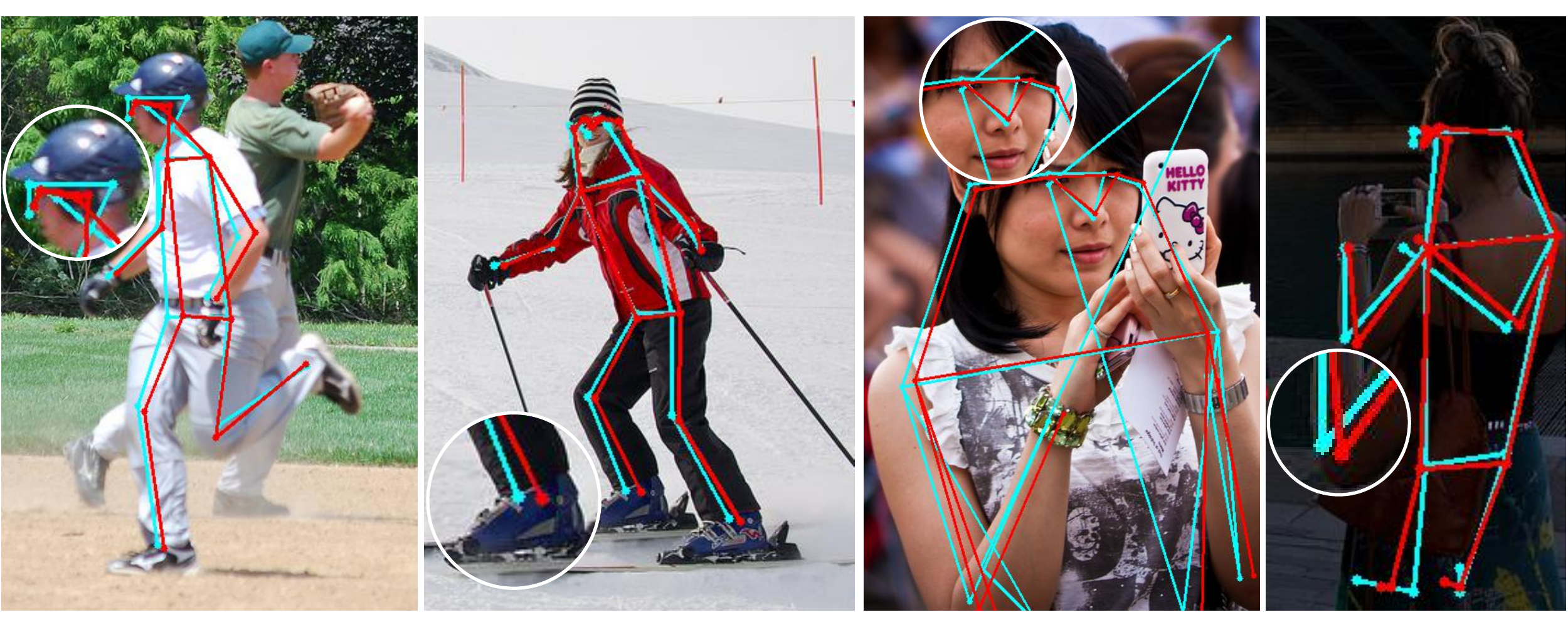}
    \caption{Qualitative evaluation of offset-based UDP \cite{UDP:huang2019devil} (red) vs. heatmap-based HRNet \cite{HRNet:0009XLW19} (cyan) on low-resolution (128$\times$96) images from COCO dataset. To some extent, UDP can locate body parts more accurately than HRNet.}
    \label{fig:illustration}
\end{figure}

In this work, we investigate the under-studied yet important low-resolution human pose estimation problem.
We start with heatmap-based methods as they are the dominant approaches for human pose estimation \cite{hourglass:NewellYD16,CPN:ChenWPZYS18}.
In particular, we choose HRNet \cite{HRNet:0009XLW19}.
Our experiments show that 
it is surprisingly sensitive to input image resolution.
One major reason is that low-resolution would exaggerate
the degree of quantization error since a fixed amount of shift against the ground-truth location means a larger error rate
due to the smaller number of pixels.
In light of this observation,
we further investigate less popular offset-based human pose methods
(\eg, UDP \cite{UDP:huang2019devil} and G-RMI \cite{GRMI:PapandreouZKTTB17})
and demonstrate that they can provide more reliable solutions
against low-resolution images.
Specifically, Figure \ref{fig:low_resolution} shows that the model performance gap between UDP and HRNet increases when the resolution of input images decreases. 
For the input size of 128$\times$96, UDP significantly improves the AP accuracy of HRNet-W48 by 8.8\%.
This indicates that offset modeling is an effective
strategy for mitigating quantization error in low-resolution data. 
Several human pose examples are given in Figure
\ref{fig:illustration} for visual examination.

Offset-based methods infer the final prediction by adding offset vectors to coarse joint coordinates.
The offset vectors are extracted from offset fields at the coarse joint coordinates.
It involves two independent tasks: 
heatmap regression and offset regression,
with the two coupled in a pixel-to-pixel correspondence manner.
During training, the activation area is often defined 
as a circle
centered at the ground-truth joint location.
The learning objective is to encourage {\em every activation pixel} to offer correct prediction.
That is, each activation pixel is treated equally important.
During testing, the peak location is first identified in
the heatmap prediction, followed by retrieving the corresponding offset value to be added for generating the final output.
This design above comes with two limitations:
(1) The training and testing processes are inconsistent in the sense that only one location is used in testing, versus all locations in training. This problem might be sourced in
the use of binary heatmaps.
(2) The offset field learning is decoupled from the coarse prediction, making the learning behaviour lose optimization focus.
Besides, both factors render most of the training efforts across all activation pixels wasted, and potentially making the model less optimal.

To overcome the aforementioned problems, we propose a novel {\em Confidence-Aware Learning} (CAL) method.
Specifically, we propose to leverage a Gaussian mask to weigh the binary heatmap.
This assigns more weights to pixels closer to ground-truth location, making training and testing more consistent and producing better coarse predictions.
Interestingly, this results in a similar learning object 
as heatmap-based methods \cite{HRNet:0009XLW19,hourglass:NewellYD16}.
Crucially, we couple the offset learning with
coarse prediction by further learning a mask to weigh 
per-pixel offset loss selectively.
This mask is learned on-the-fly from the statistics of predicted heatmaps in each individual mini-batch,
assuming it follows a Gaussian Mixture Model.

We make the following {\bf contributions} in this work:
(i) 
We investigate the low-resolution input problem in human pose estimation. It is a largely under-studied perspective in the literature, but presents practical importance in many real-world applications with poor imaging conditions.
(ii) 
We empirically investigate existing state-of-the-art methods
and reveal that offset learning is an effective approach for low-resolution human pose estimation.
(iii)
We further propose a novel {\em Confidence-Aware Learning} (CAL) method for solving the limitations of previous offset-based human pose methods.
CAL enables to selectively weigh the learning of 
heatmap and offset with respect to ground-truth and 
prediction confidence, whilst capturing the statistical 
importance of model output.
It also makes the model training and testing 
more consistent.
(iv) 
To demonstrate the effectiveness of our proposed method, 
we implement the idea of CAL with HRNet.
Extensive experiments on the COCO dataset show that our method outperforms previous state-of-the-art methods often by a large margin in the low-resolution setting,
achieving 72.3\% AP on the test-dev set when the input images are at the resolution of 128$\times$96.

The remainder of the paper is organized as follows. 
Section \ref{sec:relatedwork} gives related works on human pose estimation and image super-resolution methods; %
Section \ref{sec:revisit} revisits the key details of heatmap-based and offset-based methods.
Section \ref{sec:method} details the key components of the proposed {\em Confidence-Aware Learning} method. 
We validate our method on the commonly used COCO benchmark in Section \ref{sec:experiment}, followed by ablation study in Section \ref{sec:experiment}. Finally, we conclude this study in Section \ref{sec:conclusion}.

\section{Related work}
\label{sec:relatedwork}

\subsection{Human pose estimation}
Existing human pose estimation methods mainly fall into three categories: 1) coordinate-based methods, 2) heatmap-based methods, and 3) offset-based methods. 

{\bf \noindent
Coordinate-based methods.
}
Toshev \etal~\cite{deeppose:ToshevS14} proposed a novel framework named DeepPose which takes a 2D human body image as input and learns the normalized coordinates of body parts. 
To improve the performance of the network,
they adopted the cascaded framework to refine predictions iteratively. 
Based on DeepPose, Fan \etal~\cite{dsdn:FanZLW15} devised a dual-source solution where they fed the local image patches of body parts combined with the whole image into the network to capture holistic and local context.
Luvizon \etal~\cite{integral:LuvizonTP19} and Sun \etal~\cite{integral:SunXWLW18} attempted almost simultaneously to unify the heatmap representation and coordinate regression to establish a 
fully differentiable framework.
Nibali \etal~\cite{integral:Nibali2018} further improved the coordinate-based methods with heatmap representation by adding regularization constraints on the shape of heatmap.
To improve the performance of pose model on low-resolution images, Xu \etal~\cite{xu2020} enforced the consistency of features and outputs across resolution leveraging self-supervision and contrastive learning.

In summary, coordinate-based methods are simple and straightforward, 
but suffer from model overfitting.
Although coordinate-based methods with heatmap representation
can partially alleviate the overfitting problem in an end-to-end fashion. As the learning target of these methods, \ie, the coordinates of joints, fails to provide spatial information for pose models.

{\bf \noindent
Heatmap-based methods.
}
Heatmap-based methods have the advantage of preserving implicit body structure and can encode the coordinates of joints in the form of Gaussian distribution, which not only prevents the notorious overfitting problem but also provides a tolerance mechanism for mislabelling.
Inspired by the pose machine \cite{pm:RamakrishnaMHBS14}, Wei \etal~\cite{cpm:WeiRKS16} implemented the convolutional pose machine based on VGGNet \cite{vgg:SimonyanZ14a}
to increase the perceptive field and leveraged the multi-stage supervision to refine predictions from coarse to fine.
Newell \etal~\cite{hourglass:NewellYD16} proposed the hourglass module to extract multi-scale features by repeating the downsampling and upsampling operations and utilizing skip connections to fuse features with different resolutions.
Furthermore, they investigated the multi-scale feature representation learning by stacking several hourglass modules and adopted the intermediate supervision to improve the performance of the stacked hourglass network.
Since U-shape structure can maintain the resolution of features and possess the property of reserving semantic information in features.
Lin \etal~\cite{fpn:LinDGHHB17} presented the Feature Pyramid Network (FPN) which further improves the U-shape structure with deeply supervised information. 
Chen \etal~\cite{CPN:ChenWPZYS18} developed a Cascaded Pyramid Network (CPN) which employs a feature pyramid structure to enhance its appearance representation. In addition, they adopted an Online Hard Keypoint Mining (OHKM) strategy to correct the predictions of hard samples.
Different from Hourglass \cite{hourglass:NewellYD16} and CPN \cite{CPN:ChenWPZYS18}, Xiao \etal~\cite{simplebaseline:XiaoWW18} utilized several deconvolutional layers to obtain high-resolution feature maps and verified the effect of resolution on feature maps.

Compared with using joint coordinate directly, heatmap can provide much more effective supervision information for pose models.
However, heatmap prediction can be easily affected by the downsampling operator, which inevitably leads to quantization error.
Morevoer, the conversion from heatmap to coordinate is influenced by the resolution.
To alleviate this problem, Zhang \etal~\cite{DARK:Zhang2020} devised a universal post-processing method named Distribution-Aware Representation of Keypoint (DARK) to estimate the sub-pixel joint location. To some extent, the DARK method can mitigate the quantization error, but it relies greatly on the distribution of predicted heatmap and is affected by the resolution of heatmap heavily.
To mitigate the quantization error, Neuman \etal \cite{tiny:Neumann2018accv} converted the discrete heatmap into continuous domain by Gaussian Mixture Model for more effectively modeling the intrinsic ambiguity of tiny person instances. In contrast, in this work we leverage Gaussian distributions to learn a novel offset weighing scheme, conditioned on the displacement statistics between model predictions and the ground-truth.
Wang \etal \cite{pcnn:Wang2020eccv} proposed a two-stage elaborate framework characterized with graph pose refinement in a coarse-to-fine manner.

{\bf \noindent
Offset-based methods. 
}
To mitigate the quantization error, Papandreou \etal~\cite{GRMI:PapandreouZKTTB17} decomposed the problem of pose estimation into a part detection task and an offset field regression task.
They aggregated the heatmap predictions and the predicted offset fields by the Hough voting strategy to obtain the highly localized activation maps.
Huang \etal~\cite{UDP:huang2019devil} adopted the same formulation in \cite{GRMI:PapandreouZKTTB17}. They investigated the problem in data processing and presented an unbiased method to eliminate the error caused by flipping.

We empirically find that offset learning of these methods
are effective for low-resolution challenges.
We further address two fundamental training limitations of these 
models by confidence-aware learning 
via loss weighing at pixel level.

\subsection{Image super-resolution}
Image Super-Resolution (SR) is a technique aiming to recover a high-resolution image from the corresponding low-resolution image.
It can also be applied to human pose estimation to tackle the low-resolution problem.
Ledig \etal~\cite{sr:ledig2017photo} presented SRGAN by employing GAN to recover photo-realistic natural images and adopted the perceptual loss function to alleviate high peak signal-to-noise ratios. 
To reduce computation complexity and undesired artifacts, 
Wang \etal~\cite{sr:wang2018esrgan} designed Residual-in-Residual Dense Block (RRDB) without batch normalization to produce dense connections.
Besides, they also devised the perceptual loss in SRGAN by enhancing the supervision for brightness consistency and texture recovery. 
Zhang \etal~\cite{sr:zhang2018image} exploited the attention mechanism concurrently with Residual in Residual (RIR) structure to bypass abundant low-frequency information within low-resolution images.
To improve perceptual quality, Soh \etal~\cite{sr:soh2019natural} divided high-resolution space into three disjoint sets, \ie, blurry set, natural set and noisy set, and designed Natural Manifold Discriminator (NMD) to narrow the output space to the natural manifold.
Ma \etal~\cite{spsr:ma2020structure} proposed a Structure-Preserving Super Resolution (SPSR) method which restores high-resolution gradient maps to provide additional structure priors and designs a gradient loss to
enforce the second-order geometric structure restriction.

As a generic method, we also evaluated the benefits of image super-resolution for the proposed problem.
However, our experiments find unsatisfactory results.
Instead, we resolve the low-resolution challenge 
by developing a more effective offset learning method.

\section{Revisiting heatmap-based and offset-based methods}
\label{sec:revisit}
Given a training dataset $D=\{(\bm{x}_i,\bm{y}_i)\}_{i=1}^{N}$, human pose estimation aims to learn a mapping function $\mathcal{F}$
(with parameters $\bm{\theta}$) from an input image $\bm{x}_i$ to $K$ joint coordinates $\bm{y}_{i}=\{\bm{y}_{i}^{k}\}_{k=1}^{K}$: %
\begin{equation}
\label{eq:mapping}
\bm{y}_i = \mathcal{F}(\bm{x}_i;\bm{\theta})
\end{equation}

\subsection{Heatmap-based methods}
To provide effective supervision for pose models,
heatmap-based methods \cite{hourglass:NewellYD16,HRNet:0009XLW19,DARK:Zhang2020} usually adopt Gaussian heatmap $C_i^k(\cdot)$ to represent the joint coordinate which is formulated as:
\begin{equation}
\label{eq:gaussian heatmap}
    C_i^k(\bm{p})=exp(-\frac{\| \bm{p} - \bm{y}_i^k \|^2_2}{2 \sigma^2})
\end{equation}
where the $k$-th labelled joint coordinate is $\bm{y}_i^k$, and $\bm{p}$ enumerates all the spatial locations in the heatmap, and $\sigma$ is a fixed standard deviation of Gaussian distribution.

{\bf Model training.}
To optimize a pose model, the heatmap-based methods often utilize the $L2$ loss function defined as:
\begin{equation}
\mathcal{L}=\frac{1}{NK}\sum_{i=1}^{N}\sum_{k=1}^{K} L2(C_i^k, \hat{C}_i^k)
\end{equation}
where $\hat{C}_i^k$ is the predicted heatmap for the $k$-th joint of $i$-th person instance. $K$ and $N$ are the number of joints and training samples respectively.

{\bf Model testing.}
During model testing, the final predicted coordinate $\hat{\bm{y}}_i^k$ of $k$-th joint is obtained by $\mathop{\arg\max}$ operation as:
\begin{equation}
\hat{\bm{y}}_i^k = \mathop{\arg\max}_{\bm{p}} (\hat{C}_i^k)
\end{equation}

{\bf
Limitations.}
The downsampling operations, which are applied for reducing computational overhead,
leads to a need for resolution recovery so that the final prediction can reside
in the input coordinate space.
This will bring about quantization errors and finally affect the model performance negatively \cite{DARK:Zhang2020}.
Unfortunately, this error would exaggerate when low-resolution images are presented
as a fixed shift takes a larger proportion of the whole image.  

\subsection{Offset-based methods}
\begin{figure}
    \centering
    \includegraphics[width=4.7in]{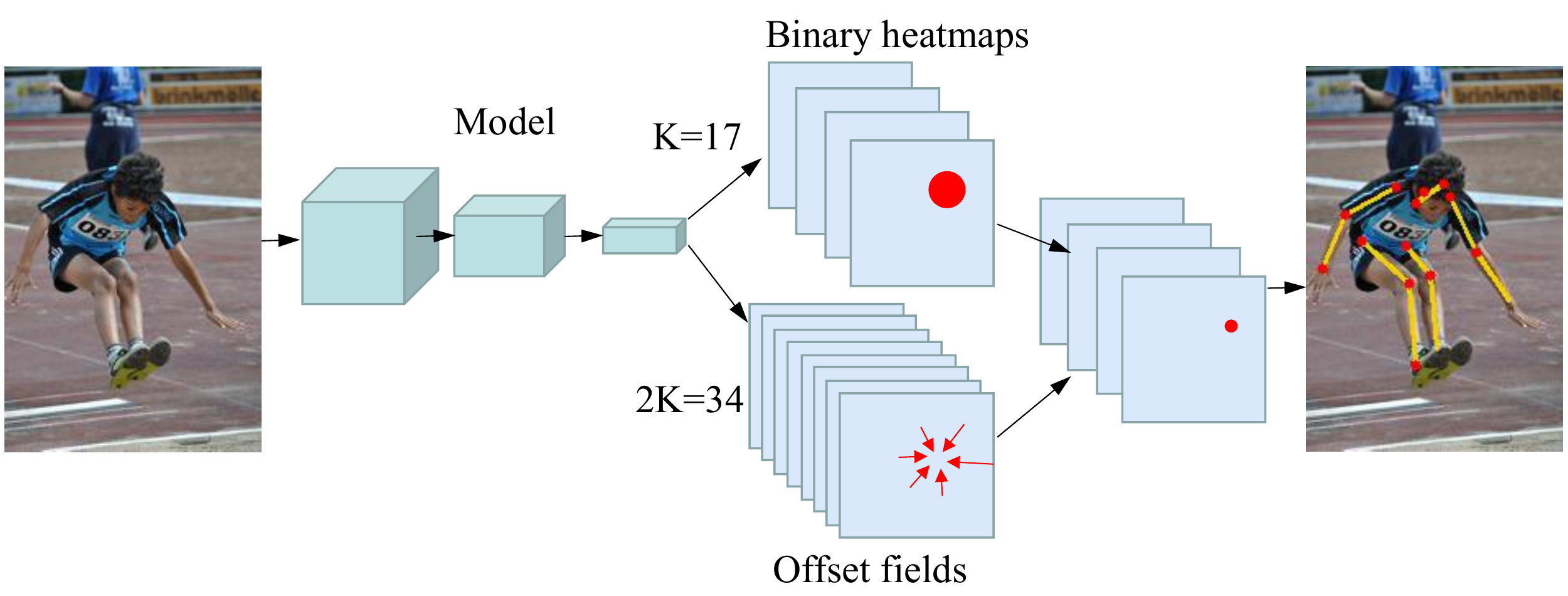}
    \caption{The pipline of offset-based methods.}
    \label{fig:grmi}
\end{figure}

Offset-based methods can alleviate this aforementioned limitation with heatmap-based methods by decomposing the pose estimation problem into 
two parts (see Figure \ref{fig:grmi}): a part detection task and an offset field regression task, \ie, the estimation of the
coarse position $\hat{\bm{c}}_{i}^{k}$ and the offset vector $\hat{\bm{o}}_{i}^{k}$ in Eq.\eqref{eq:twotasks}.
\begin{equation}
\label{eq:twotasks}
\hat{\bm{y}}_{i}^{k} = \hat{\bm{c}}_{i}^{k} + \hat{\bm{o}}_{i}^{k}
\end{equation}

{\bf Model training.}
The ground truth for training offset-based human pose models is represented by a binary heatmap $B_i^k$ and an offset field $O_i^k$ jointly as:
\begin{equation}
B_i^k(\bm{p})=
\begin{cases}
1 & ||\bm{p}-\bm{y}_i^k||_2 \le R \\
0 & otherwise\\
\end{cases}
\end{equation}

\begin{equation}
O_i^k(\bm{p})=
\begin{cases}
\bm{y}_i^k-\bm{p} & \|\bm{p}-\bm{y}_i^k\|_2 \le R \\
\bm{0} & otherwise\\
\end{cases}
\end{equation}
where the {\em positive} area is a $R$-radius circle
centered at the $k$-th labelled joint coordinate $\bm{y}_i^k$.
The offset field uses $\bm{y}_i^k$ as the target point.  
The two supervision signals have the same spatial resolution
with point-to-point correspondence.

In training, the cross-entropy $CE(\cdot)$ and smooth $L1$ $SmoothL1(\cdot)$ loss functions are adopted on the binary heatmap and offset field supervision, respectively.
Formally, the objective loss function is defined as:
\begin{equation}
\label{eq:offsetloss}
\mathcal{L}= \frac{1}{NK}\sum_{i=1}^{N}\sum_{k=1}^{K} [CE(B_i^k, \hat{B}_i^k) + \alpha \cdot SmoothL1(O_i^k, \hat{O}_i^k)]
\end{equation}
where $\alpha$ is a hyper-parameter to balance the two losses,
$\hat{B}_i^k$ and $\hat{O}_i^k$ denote the predicted binary heatmap and predicted offset field.

{\bf Model testing.}
During model testing, the predicted binary heatmap $\hat{B}_i^k$ and predicted offset field $\hat{O}_i^k$ are used together to obtain the final predicted coordinate $\hat{\bm{y}}_i^k$ of $k$-th joint as:
\begin{equation}
\label{eq:decode}
    \hat{\bm{y}}_i^k = \mathop{\arg\max}_{\bm{p}} (\hat{B}_i^k) + \hat{O}_i^k\Big(\mathop{\arg\max}_{\bm{p}} (\hat{B}_i^k)\Big)
\end{equation}

Concretely, the most confident location in heatmap 
is first taken as the coarse prediction (\ie, the first term) which is further
fine-tuned by the corresponding offset vector (\ie, the second term).

\begin{figure}
\centering
\includegraphics[width=4.6 in]{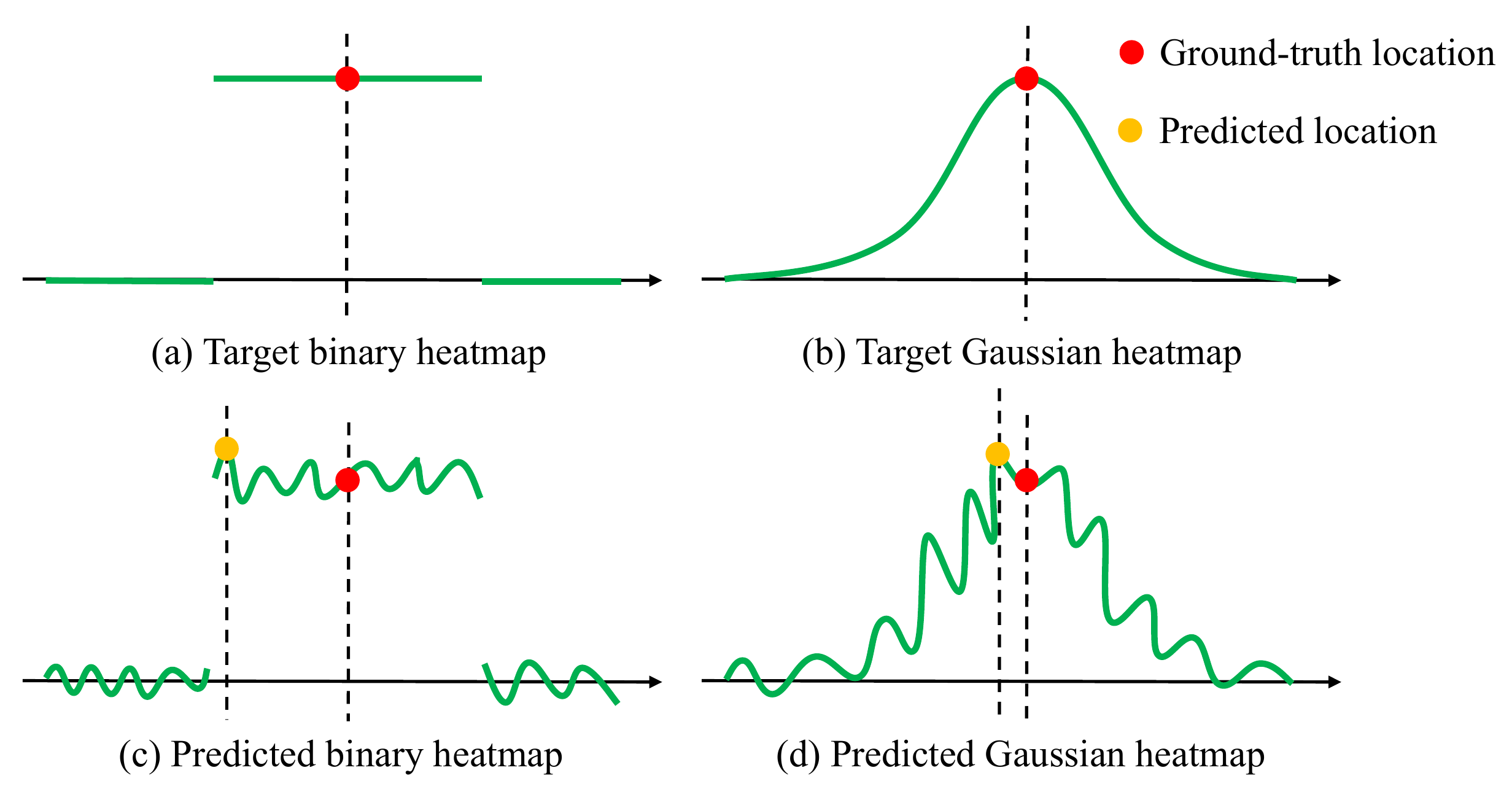}
\caption{The lateral view of binary heatmap and Gaussian heatmap. The predicted binary heatmap presents multiple peaks with the prediction highly deviated from the ground-truth location, due to higher learning difficulty and ambiguity. In contrast, the predicted Gaussian heatmap can more accurately capture the ground-truth joint location.}
\label{fig:heatmap}
\end{figure}

{\bf Limitations.}
There are a couple of limitations with existing offset-based methods.
(i) They adopt a binary heatmap as training supervision where all pixels around the ground truth within a radius $R$ circle are set to $1$, which is inconsistent with the decoding process via $\mathop{\arg\max}$ operation.
Empirically, we find this supervision will lead to 
coarse prediction with large deviation from the ground truth 
due to the hardness of model fitting,
as illustrated in Figure \ref{fig:heatmap}(c).
(ii) Learning the offset field is decoupled from the coarse prediction. This makes the learning behaviour lose optimization 
focus and hence potentially less optimal.

\section{Confidence-Aware Learning}
\label{sec:method}
\subsection{Gaussian heatmap weighing}
Existing offset-based methods, such as UDP \cite{UDP:huang2019devil}, employ a binary heatmap to represent the activation region of a body joint.
The confidence values within the activation circle area are all $1$.
Such a learning target fails to provide 
spatial relationships with respect to the ground-truth joint location.
Specifically, it cannot reflect the fact that the confidence values of pixels closer to ground-truth should be larger than those further away.  

To solve this issue above, we scale the binary heatmap by a Gaussian distribution $C_i^k(\bm{p})$ 
in order to encode the desired confidence distribution 
as: 
\begin{equation}
G_i^k(\bm{p})=B_i^k(\bm{p}) \cdot C_i^k(\bm{p})
\end{equation}

Using binary heatmap as training target, the multiple peaks in the predicted heatmap make the $\mathop{\arg\max}$ operation produce highly deviated coordinates (as shown in Figure \ref{fig:heatmap}(c)).
Our proposed Gaussian weighing strategy
addresses this issue by introducing proper confidence values
in model training.
As a result, a more accurate coarse prediction can be yielded.

\subsection{Gaussian offset weighing}
The offset fields in offset-based methods aim to compensate for the displacements between the coarse predictions and ground-truth positions.
However, existing offset learning methods \cite{GRMI:PapandreouZKTTB17,UDP:huang2019devil} 
identically treat all activation pixels and ignore
the coarse prediction of heatmap, \ie, decoupled.
On the contrary, only the offset at the location of coarse prediction
is useful during testing.
This leads to a mismatch between training and testing.

\begin{figure}[h]
    \centering
    \includegraphics[width=4.8in]{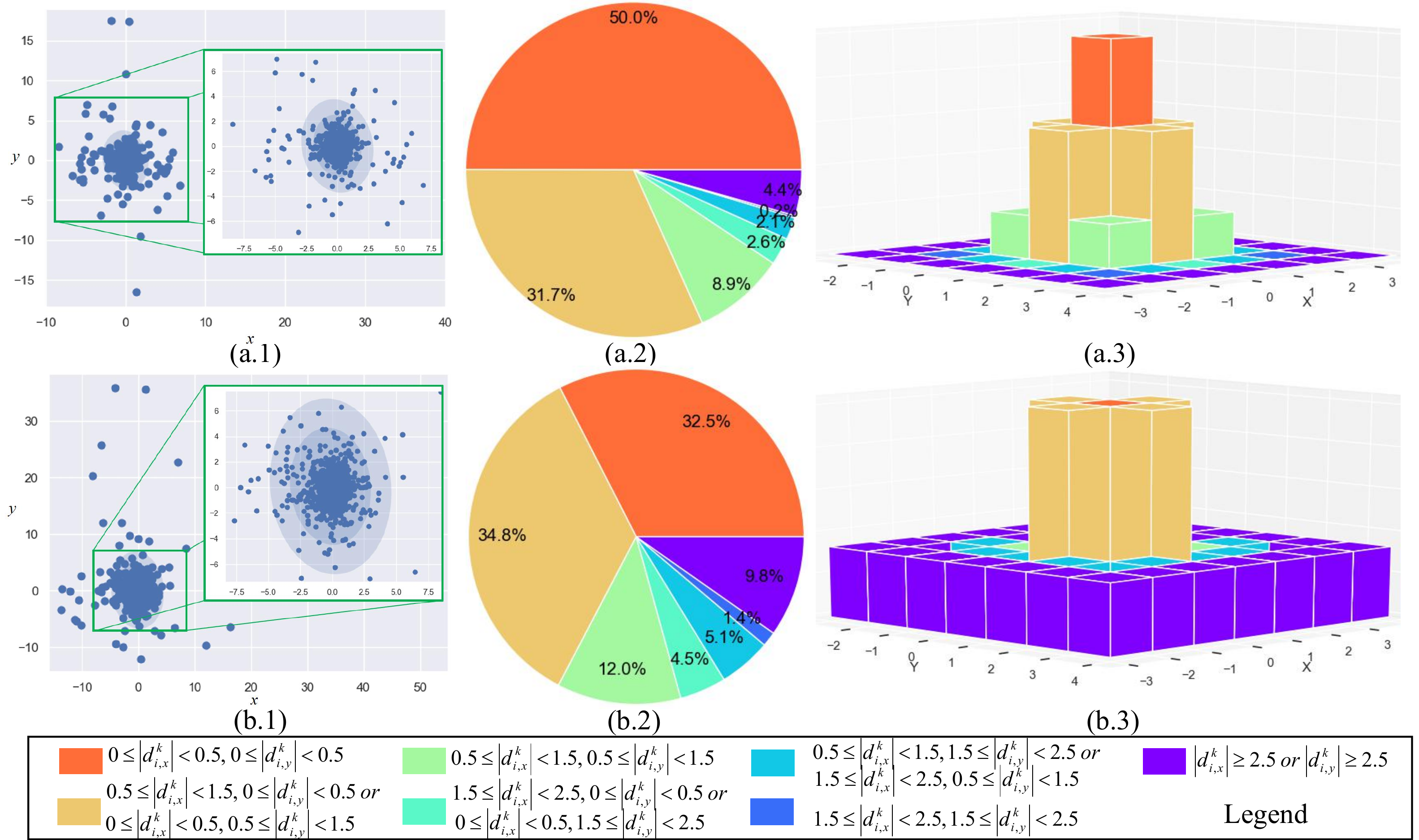}
    \caption{The distribution of relative displacements from the coarse predictions to ground-truth positions.
    We divide the entire range of displacement into a series of intervals and then count how many displacements fall into each interval. The bin (bucket) is shown in the legend.
    We compare the distributions of two image resolutions 
    (a) 128$\times$96 and (b) 256$\times$192,
    using the HRNet-W32 based UDP pose model.
    First column: the scatter of relative displacements.
    Second column: the share of each displacement bin. 
    Third column: the corresponding 2D histogram in the space.
    }
    \label{fig:distribution}
\end{figure}

To overcome this problem,
we consistently introduce Gaussian offset weighing.
Instead of manually setting a weight masking as in heatmap regression,
we propose to learn the weight parameters as
the offset, when conditioned on the coarse prediction, 
presents more complex patterns.
Concretely, we first infer the distribution of displacements between the coarse predictions and ground-truth positions, and then apply
them to coarse prediction as offset loss weights.

More specifically, we investigate the offset error distribution 
by collecting relative displacements: 
\begin{equation}
    \mathbb{S}=\{ \bm{d}_i^k | \bm{d}_i^k =\hat{\bm{y}}_i^k - \bm{y}_i^k \}
\end{equation}
from the coarse predictions \{$\hat{\bm{y}}_i^k$\} to ground-truth positions \{$\bm{y}_i^k$\}.
Figure \ref{fig:distribution} gives the distribution of these displacements.
It is shown that they are largely in Gaussian distributions.

To quantify these displacements, we therefore assume and learn 
Gaussian Mixture Model (GMM) for offset weighing.
This is because GMM can be efficiently inferred with solid theories (taking around 12\% of training time in our experiments).
Specifically, with the $\mathop{\arg\max}$ operation all the positions $\hat{\mathbb{Y}}=\{ \hat{\bm{y}}_i^k | \hat{\bm{y}}_i^k = \mathop{\arg\max}_{\bm{p}} (\hat{G}_i^k) \}$ with maximum values in a mini-batch of predicted heatmaps are first obtained.
The collection of relative displacements $\mathbb{S}$ 
is then fed into an initialized Gaussian Mixture Model $H(\cdot;\bm{\theta})$ whose parameter $\bm{\theta}$ is optimized by the Expectation-Maximization (EM) algorithm \cite{Dempster1977Maximum}.

\begin{equation}
    \hat{\bm{\theta}} = \mathop{\arg\max}_{\bm{\theta}} \log( H (\mathbb{S}; \bm{\theta}) )
\end{equation}

\begin{figure}[h]
    \centering
    \includegraphics[width=4.2in]{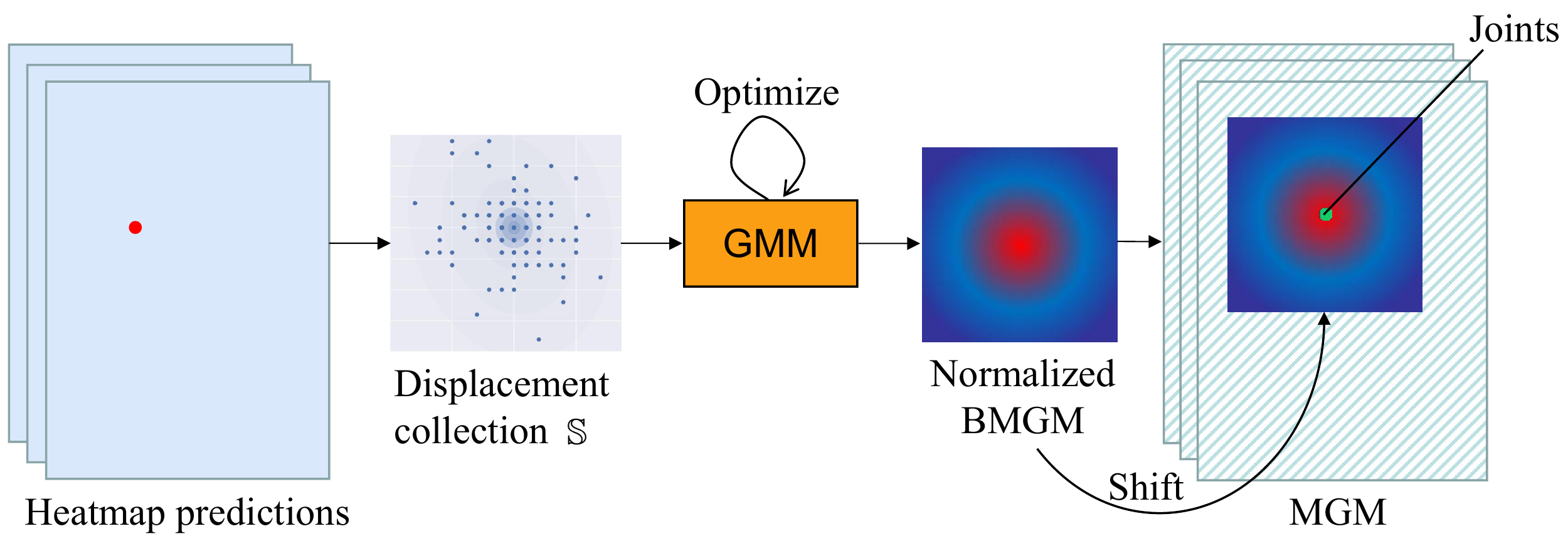}
    \caption{Generation of Mixed Gaussian Masks (MGMs) in Gaussian offset weighing.
    A mini-batch of relative displacements $\mathbb{S}$ are first collected from the predicted heatmaps with respect to the ground-truth annotation. They are then fed into the Gaussian Mixture Model to obtain the Initial Mixed Gaussian Mask (IMGM).
    Finally, we normalize IMGM and shift it according to the locations of joints to generate the final MGMs.
    }
    \label{fig:mgm}
\end{figure}

Due to that the heatmap is discrete, we then sample the learned Gaussian Mixture Model $H(\cdot;\hat{\theta})$ over a regular grid of $ \mathbb{Z}=\{ (-R,-R),(-R+1, -R), \dots ,(R-1, R), (R,R) \}$ to obtain the Initial Mixed Gaussian Mask (IMGM), denoted as $\bm{M}$. 
IMGM is further normalized. This summarizes the statistics over a mini-batch of predicted heatmaps. 

Eventually, the center of IMGM $\bm{M}$ is shifted to the location of each joint to incorporate spatial information of each joint $\bm{y}_{i}^{k}$ to generate a series of MGMs $\{m_i^k\}$ as:
\begin{equation}
    m_i^k = shift(norm(\bm{M}), \bm{y}_{i}^{k})
\end{equation}

It is worthy to note that the generated MGMs $\{m_i^k\}$ not only capture the offset error statistics of predicted heatmaps, but also contain the spatial information from each joint. During training, with the proposed MGM the gradients of offset vectors closer to the ground-truth joints are assigned with more weights. Functionally, MGM serves as the bridge between the offset maps and heatmaps in model optimization. This design is useful as it enables the optimization of the two components compatible to make more effective model training.
In contrast, all previous offset-based methods ignore their mutual relationship in model optimization.
This is shown to be critical in improving the accuracy of the offset vectors.
We summarize the procedure of generating MGMs in Algorithm \ref{algo}.

\begin{algorithm}[h]
\caption{The generation of Mixed Gaussian Masks}
\label{algo}
\textbf{Initialize:} The parameter $\bm{\theta}$ of Gaussian Mixture Model $H(\cdot;\theta)$. \\
\textbf{Input:} A mini-batch of predicted heatmaps $ \{ \hat{G}_i^k \}$, corresponding ground-truth $ \{ \bm{y}_i^k \}$, number of samples in a mini-batch $N_b$, number of joints $K$, regular grid $ \mathbb{Z}$. \\
\textbf{Output:} MGMs $\{ m_i^k \}$
\begin{algorithmic}[1]
\State{$\mathbb{\hat{Y}}\leftarrow\{\hat{\bm{y}}_i^k | \hat{\bm{y}}_i^k= \mathop{\arg\max}_{p} (\hat{G}_i^k) \}$}
\State{$\mathbb{S} \leftarrow \{ \bm{d}_i^k | \bm{d}_i^k =\hat{\bm{y}}_i^k - \bm{y}_i^k \}$}
\State{$\hat{\bm{\theta}} \leftarrow \mathop{\arg\max}_{\bm{\theta}} log( H (\mathbb{S} ; \bm{\theta}) )$ }
\State{$\bm{M} \leftarrow H(\mathbb{Z}; \hat{\bm{\theta}})$ }
\State{$\bm{M} \leftarrow \frac{\bm{M} - \min(\bm{M})}{\max(\bm{M})-\min(\bm{M})}$}
\For{i=$1,\dots,N_b$}
    \For{k=$1,\dots,K$}
        \State{$m_i^k \leftarrow shift(\bm{M}, y_i^k) $}
    \EndFor
\EndFor
\State \Return{$\{ m_i^k\}$}  
\end{algorithmic}
\end{algorithm}

\subsection{Model training and testing}

Since the MGMs depend on the predicted heatmaps, 
we propose a two-stage training process for our method
where MGM is introduced after heatmap prediction 
becomes sufficiently strong. 

In the first stage, we train the model without MGMs.
The loss function $\mathcal{L}_{s1}$ is formulated as:

\begin{equation}
    \mathcal{L}_{s1}= \frac{1}{NK}\sum_{i=1}^{N}\sum_{k=1}^{K} [L2(G_i^k, \hat{G}_i^k) + \alpha G_i^k L1(O_i^k, \hat{O}_i^k)]
\end{equation}
where we adopt $L2$ for the heatmap regression. Instead of Smooth $L1$ as used in previous offset-based methods \cite{UDP:huang2019devil,GRMI:PapandreouZKTTB17}, we select $L1$ for offset field regression. 
This is because Smooth $L1$ may distort the proposed weight assignment. %

After the predicted heatmaps become strong with sufficient spatial information for offset fields, 
we utilize the predicted heatmaps to generate MGMs.
The loss function $\mathcal{L}_{s2}$ used in the second stage is defined as: 
\begin{equation}
\label{eq:mgm}
    \mathcal{L}_{s2}= \frac{1}{NK}\sum_{i=1}^{N}\sum_{k=1}^{K} [L2(G_i^k, \hat{G}_i^k) + \alpha m_i^k L1(O_i^k, \hat{O}_i^k)]
\end{equation}

During testing, as existing offset-based methods \cite{UDP:huang2019devil,GRMI:PapandreouZKTTB17}, we obtain the final predictions from predicted heatmaps and offset fields by Eq.\eqref{eq:decode}. 
To give an intuitive and simple understanding of the inference procedure, We summarize it in Algorithm \ref{algo:2}. 

\begin{algorithm}[h]
\caption{Model inference at the testing phase}
\label{algo:2}
\textbf{Initialize:} A trained pose model $PM(\cdot)$, number of samples in a mini-batch $N_b$, number of joints $K$. \\
\textbf{Input:} An mini-batch input image $\mathbb{I}$ \\
\textbf{Output:} Coordinate of predicted joints $\mathbb{\hat{Y}}=\{ \hat{y}_i^k \}$
\begin{algorithmic}[1]
\State{$\{ \hat{G}_i^k,\hat{O}_i^k \} \leftarrow PM(\mathbb{I})$}
\For{i=$1,\dots,N_b$}
    \For{k=$1,\dots,K$}
        \State{$\hat{y}_i^k \leftarrow \mathop{\arg\max} (\hat{G}_i^k) $}
        \State{$\hat{y}_i^k \leftarrow \hat{O}_i^k(\hat{y}_i^k) + \hat{y}_i^k$ }
    \EndFor
\EndFor
\State \Return{$\mathbb{\hat{Y}}=\{ \hat{y}_i^k\}$}  
\end{algorithmic}
\end{algorithm}

\section{Experiments}
\label{sec:experiment}
\subsection{Dataset and evaluation metric}
The COCO dataset \cite{dataset:coco} is a large-scale benchmark for object detection, keypoint detection, and instance segmentation. It is collected from various daily activities and includes 287K images and 1.5 million object instances. The dataset can be split into four parts: 118K images for training, 5K images for validation, 41K images for testing and 123K unlabelled images. Now it has become the most popular benchmark for pose estimation. In this paper, we trained and evaluated our proposed method on the COCO dataset. We adopted the Object Keypoint Similarity (OKS) to measure the similarity between predictions and ground truths, and took OKS-based Average Precision (AP) and Average Recall (AR) as the evaluation metrics to indicate the prediction accuracy. 
The definition of OKS is illustrated as follows: 
\begin{equation}\label{eq_oks}
\text{OKS}=\frac{\sum_i \exp(\frac{-d^2_i}{2S^2\sigma^2_i}) \delta(v_i >0)}{\sum_i \delta(v_i >0) }
\end{equation} 
where $d$ represents the Euclidean distance between the predicted location of joint and ground-truth location of joint, the indicator function $\delta(v_i > 0)$ returns $1$ if the $i$-th joint is labelled, and scale factor $S$ is the square root of the bounding box area. The variance $\sigma_i$ of each type of joint is different.

\subsection{Implementation details}
To compare with the baseline fairly, the configuration of our method strictly follows UDP \cite{UDP:huang2019devil}. The common data augmentations, \ie, random rotation with $[-45^{\circ}, 45^{\circ}]$, random flipping with a half probability, random scaling $([0.65, 1.35])$, are used in our method. 
Besides, Adam optimizer is employed with an initialized learning rate $1e-3$ which decays to $1e-4$ and $1e-5$ at the 170-th and 200-th epoch respectively. 
The training process contains two stages containing 210 epochs and 40 epochs respectively.
The number of components in the Gaussian Mixture Model is set to 1 for low-resolution input (64$\times$48 and 128$\times$96) and 2 for high-resolution input (256$\times$192 and 384$\times$288).
In inference, we adopt the same human instance detector as UDP \cite{UDP:huang2019devil}, which achieves 65.1\% AP on the COCO val set. The performance of single model with flipping is reported.

\subsection{Comparison with the state-of-the-art on low-resolution pose estimation} 
We compare our method with three state-of-the-art methods under the low-resolution setting, \ie, two heatmap-based methods (SimpleBaseline \cite{simplebaseline:XiaoWW18} and HRNet \cite{HRNet:0009XLW19}) and one offset-based method (UDP).
The same resolution of 128$\times$96 is used for both model training and testing.

Table \ref{tab:STOA_COCO_test} shows that CAL with HRNet-W48 achieves 72.3\% AP, outperforming all the state-of-the-art methods on the COCO test-dev set. 
(i) Compared to the best heatmap-based method, CAL achieves 11.1\% (72.3-61.2) AP improvement.
(ii) Compared to the offset-based method (UDP), it obtains 2.6\% (72.3-69.7) AP gain.
These results indicate that CAL has a great advantage compared with both heatmap-based methods and offset-based method in low-resolution setting.

\begin{table*}[htb]
	\setlength{\tabcolsep}{0.1cm}
	\begin{center}
		\resizebox{\columnwidth}{!}{%
		\begin{tabular}{ l | c | c | c | c | c | c | c | c | c | c  }
			\hline
			Method & Backbone & Input size & \#Params & GFLOPs & $AP$ & $AP^{50}$ & $AP^{75}$ & $AP^{M}$ & $AP^{L}$ & $AR$ \\
			\hline
			SimpleBaseline\cite{simplebaseline:XiaoWW18} & ResNet-50 & $128\times96$ & - & -
			& 60.3 & 88.1 & 67.8 & 58.6 & 62.8 & 64.4
			\\
			SimpleBaseline\cite{simplebaseline:XiaoWW18} & ResNet-101 & $128\times96$ & - & - 
			& 59.7 & 88.2 & 67.5 & 58.6 & 61.7 & 64.1
			\\
			SimpleBaseline\cite{simplebaseline:XiaoWW18} & ResNet-152 & $128\times96$ & - & - 
			& 61.1 & 88.2 & 69.7 & 59.9 & 63.8 & 65.5
			\\
			HRNet \cite{HRNet:0009XLW19} & HRNet-W32 & $128 \times 96$ & 28.5M & 1.8
			& 60.0 & 90.2 & 70.1 & 57.4 & 64.9 & 67.2
			\\
			HRNet \cite{HRNet:0009XLW19} & HRNet-W48 & $128 \times 96$ & 63.6M & 3.7
			& 61.2 & 90.6 & 71.4 & 58.6 & 66.3 & 68.4
			\\
			UDP \cite{UDP:huang2019devil} & HRNet-W32 & $128 \times 96$ & 28.6M & 1.8
			& 67.9 & 90.6 & 76.2 & 65.9 & 72.3 & 74.3
			\\
			UDP \cite{UDP:huang2019devil} & HRNet-W48 & $128 \times 96$ & 63.7M & 3.7
			& 69.7 & 90.9 & 78.1 & 67.6 & 74.2 & 75.9
			\\
			\hline
			CAL & HRNet-W32 & $128 \times 96$ & 49.3M & 2.5
			& \bf{71.4} & \bf{91.3} & \bf{79.6} & \bf{69.2} & \bf{76.1} & \bf{77.3}
			\\
			CAL & HRNet-W48 & $128 \times 96$ & 110.3M & 5.3
			& \bf{72.3} & \bf{91.6} & \bf{80.5} & \bf{69.9} & \bf{77.2} & \bf{78.0}
			\\
			\hline
		\end{tabular}
		}
	\end{center}
	\caption{
		Comparison with the state-of-the-art human pose estimation methods in the low-resolution setting on the COCO test-dev set. 
	}
	\label{tab:STOA_COCO_test}
	\vskip -0.2cm
\end{table*}

\subsection{How image super-resolution helps?}
We investigate the effect of image super-resolution by 
comparing CAL with state-of-the-art human pose methods \cite{HRNet:0009XLW19,UDP:huang2019devil} 
with extra assistance from image super-resolution as pre-processing.
Specifically, the low-resolution (64$\times$48) images are first resolved by the SPSR model \cite{spsr:ma2020structure} to 256$\times$192 before human pose estimation.
Although Table \ref{tab:srmodel} shows that 
SPSR can boost the performance of pose models on low-resolution images,
we find that CAL outperforms clearly all the competitors by a large margin.
This demonstrates the superiority of CAL for low-resolution pose estimation over image super-resolution pre-processing.

\begin{table}[ht]
    \centering
    \scalebox{0.8}{
    \begin{tabular}{ c | c | c | c |c | c | c | c}
        \hline
        Method & Backbone & $AP$ & $AP^{50}$ & $AP^{75}$ & $AP^{M}$ & $AP^{L}$ & $AR$ \\
        \hline \hline
        HRNet & HRNet-W32 & 29.7 & 75.7 & 13.1 & 29.3 & 30.7 & 37.3 \\
        HRNet & HRNet-W48 & 32.4 & 78.3 & 16.2 & 31.5 & 34.0 & 39.3 \\
        UDP   & HRNet-W32 & 47.4 & 80.5 & 50.6 & 47.7 & 47.7 & 53.8 \\
        UDP   & HRNet-W48 & 51.0 & 82.6 & 55.2 & 51.4 & 51.0 & 57.3 \\  
        \hline
        HRNet+SPSR & HRNet-W32 & 50.0 & 81.6 & 53.6 & 53.6 & 46.0 & 55.3 \\
        HRNet+SPSR & HRNet-W48 & 51.2 & 82.8 & 55.5 & 54.6 & 47.7 & 56.4 \\
        UDP+SPSR  & HRNet-W32 & 52.5 & 80.8 & 57.7 & 56.2 & 48.3 & 57.4 \\
        UDP+SPSR & HRNet-W48 & 54.1 & 82.4 & 59.0 & 56.9 & 50.6 & 59.4 \\
        \hline
        CAL & HRNet-W32  & \bf{58.4} & \bf{86.6} & \bf{65.1} & \bf{57.3} & \bf{60.5} & \bf{64.8} \\
        CAL & HRNet-W48   & \bf{61.5} & \bf{88.1} & \bf{68.7} & \bf{60.7} & \bf{63.5} & \bf{66.3} \\
        \hline
    \end{tabular}
    }%
    \caption{Effect of image super-resolution on the COCO valid set. 
    }
    \label{tab:srmodel}
\end{table}

\subsection{Ablation study}
In this section, we evaluate CAL based on the HRNet-W32 backbone. To demonstrate the advantages of CAL, low-resolution images with the input size 128$\times$96 are taken as input to perform the ablation study.
We explore the effects of each component in CAL
and evaluate the effectiveness of 
the proposed loss function and the corresponding two-stage training strategy.

\subsubsection{Effect of Gaussian offset weighing}
We compared our proposed MGM in Gaussian offset weighing with the binary heatmap to demonstrate its effectiveness.
Specifically, we replaced the mask $m_i^k$ generated by our MGM (Eq.\eqref{eq:mgm}) with the binary heatmap.
Table \ref{tab:ommask} shows that compared with the binary heatmap, MGM can bring about 1.0\% (74.3-73.3) AP improvement. 
This verifies that the MGM enables better learning offset fields. 

\begin{table}[ht]
    \centering
    \scalebox{0.8}{
    \begin{tabular}{ c | c | c |c | c | c | c}
        \hline
        Mask Type & $AP$ & $AP^{50}$ & $AP^{75}$ & $AP^{M}$ & $AP^{L}$ & $AR$ \\
        \hline \hline
        Binary heatmap & 73.3 & 92.5 & 81.5 & 70.9 & 76.9 & 76.3 \\
        MGM & \bf{74.3} & \bf{92.6} & \bf{81.5} & \bf{72.1} & \bf{77.5} & \bf{77.2} \\
        \hline
    \end{tabular}}
    \caption{Effect of Gaussian offset weighing on the COCO valid set.}
    \label{tab:ommask}
\end{table}

\subsubsection{Effect of the component number in GMM}
To test the effect of the component number in GMM, we evaluated the HRNet-W32 based CAL with different resolution configurations. CAL with one component in GMM achieved 74.3\% AP for the low-resolution case (128$\times$96) and CAL with two components obtained 78.6\% AP for the high-resolution case (256$\times$192). The results indicate that the optimal component number of GMM is 1 for low-resolution input and 2 for high resolution input. This is because in low-resolution case, the distribution of relative displacement is similar to the Gaussian distribution, hence one GMM component is sufficient. Whilst applying more GMM components might skew the MGM and finally degrade the performance of pose model. On the contrary, in the high-resolution case the distribution of relative displacements is more sparse and distributed. As a result, one GMM component is not able to accurately simulate the distribution of relative displacements. We find that using two GMM components is a good selection to prevent the false positive modes in GMM from generating the skewed MGM.

\begin{table}[ht]
    \centering
    \scalebox{0.8}{
    \begin{tabular}{ c | c | c | c |c | c | c | c}
        \hline
        Resolution & Components & $AP$ & $AP^{50}$ & $AP^{75}$ & $AP^{M}$ & $AP^{L}$ & $AR$ \\
        \hline \hline
        \multirow{3}{*}{128$\times$96}  & 1 & \bf{74.3} & \bf{92.6} & \bf{81.5} & \bf{72.1} & \bf{77.5} & \bf{77.2} \\
                                        & 2 & 73.6 & 92.5 & 81.4 & 71.2 & 77.2 & 76.6 \\
                                        & 3 & 72.8 & 92.5 & 80.5 & 70.9 & 76.2 & 76.0 \\
        \hline
        \multirow{3}{*}{256$\times$192} & 1 & 78.5 & 93.6 & 84.8 & \bf{75.9} & 83.0 & 81.2 \\
                                        & 2 & \bf{78.6} & \bf{93.6} & \bf{85.9} & 75.6 & \bf{83.1} & \bf{81.0} \\
                                        & 3 & 78.3 & 93.6 & 84.9 & 75.5 & 83.0 & 80.9 \\
        \hline
    \end{tabular}}
    \caption{Effect of the component number in GMM on the COCO valid set.}
    \label{tab:component}
\end{table}

\subsubsection{Effect of Gaussian heatmap weighing}
The learning target in UDP \cite{UDP:huang2019devil} and G-RMI \cite{GRMI:PapandreouZKTTB17}
is the detection map with binary values, \ie, binary heatmap.
To demonstrate the superiority of Gaussian heatmap weighing in encoding joint locations,
we compared the performance of two models trained using 
binary and Gaussian heatmaps respectively.
Table \ref{tab:hmtype} shows that Gaussian heatmap can boost the performance of the pose model with 2.0\% (74.3-72.3) AP improvement over binary heatmap
\cite{UDP:huang2019devil,GRMI:PapandreouZKTTB17}.

\begin{table}[ht]
    \centering
    \scalebox{0.8}{
        \begin{tabular}{ c | c | c |c | c | c | c}
            \hline
            Heatmap Type & $AP$ & $AP^{50}$ & $AP^{75}$ & $AP^{M}$ & $AP^{L}$ & $AR$ \\
            \hline \hline
            Binary   & 72.3 & 91.5 & 80.3 & 70.5 & 75.7 & 75.5\\
            Gaussian & \bf{74.3} & \bf{92.6} & \bf{81.5} & \bf{72.1} & \bf{77.5} & \bf{77.2} \\
            \hline
        \end{tabular}
    }
    \caption{Effect of Gaussian heatmap weighing on the COCO valid set.}
    \label{tab:hmtype}
\end{table}

\subsubsection{Effect of loss function for offset field}
To find the suitable loss function for offset field, we conducted experiments where three commonly used loss functions, $L1$, Smooth $L1$ and $L2$ are adopted to train the pose model.
Table \ref{tab:losstype} shows that the pose model trained with the $L1$ loss performs better than the other models, obtaining 1.5\% (74.3-72.8) AP improvement.
Although Smooth $L1$ loss is the best choice for UDP and G-RMI,
$L1$ loss is shown to be more suitable for our method.
A plausible reason is that Smooth $L1$ loss might destroy the MGM masking for offset fields and finally hurt the model performance. 

\begin{table}[h]
    \centering
    \scalebox{0.8}{
        \begin{tabular}{ c | c | c |c | c | c | c}
            \hline
            Loss Type & $AP$ & $AP^{50}$ & $AP^{75}$ & $AP^{M}$ & $AP^{L}$ & $AR$ \\
            \hline \hline
            $L2$        & 72.8 & 91.6 & 80.5 & 70.6 & 76.5 & 75.9 \\
            Smooth $L1$ & 72.8 & 92.5 & 80.5 & 70.7 & 76.2 & 76.0 \\
            \hline
            $L1$        & \bf{74.3} & \bf{92.6} & \bf{81.5} & \bf{72.1} & \bf{77.5} & \bf{77.2} \\
            \hline
        \end{tabular}
    }%
    \caption{Effect of loss function for offset field on the COCO valid set.}
    \label{tab:losstype}
\end{table}

\subsubsection{Effect of two-stage training strategy}
In Section \ref{sec:method}, we introduce a two-stage training strategy. 
To evaluate its effect, we compare it with the one-stage training strategy where we adopt MGM (Eq.\eqref{eq:mgm}) to guide the learning of offset fields from the beginning to the end of training.
As shown in Table \ref{tab:strategy}, the 0.6\% (74.3-73.7) AP gain demonstrates the effectiveness of our two-stage training strategy.

\begin{table}[ht]
    \centering
    \scalebox{0.8}{
    \begin{tabular}{ c | c | c |c | c | c | c}
        \hline
        Training Strategy & $AP$ & $AP^{50}$ & $AP^{75}$ & $AP^{M}$ & $AP^{L}$ & $AR$ \\
        \hline \hline
        One-stage & 73.7 & 92.5 &\bf 81.5 & 71.3 & 77.1 & 76.6 \\
        \hline
        Two-stage & \bf{74.3} & \bf{92.6} & \bf{81.5} & \bf{72.1} & \bf{77.5} & \bf{77.2} \\
        \hline
    \end{tabular}
    }
    \caption{Effect of two-stage training strategy on the COCO valid set.}
    \label{tab:strategy}
\end{table}

As shown in the ablation study above, our CAL method benefits from the proposed Gaussian heatmap weighing, the loss function and training strategy to outperform all the competitors. Among these, Gaussian heatmap weighing contributes the most to the final results as compared to the other components. Besides, MGM and $L1$ loss function are supposed to work together, since the latter helps avoid the distortion of MGM weight assignment. Finally, we find that our two-stage training strategy effectively avoids the volatility of heatmap predictions, bringing about 0.6\% AP gain.

\subsubsection{Effect of low-resolution input}
To prove the superiority of proposed CAL,
we tested two resolutions 64$\times$48 and 128$\times$96 on HRNet \cite{HRNet:0009XLW19}, UDP \cite{UDP:huang2019devil} and DARK \cite{DARK:Zhang2020}.
The results in Table \ref{tab:resolution} show that:
(i) For the resolution of 64$\times$48, CAL achieves 11.0\% (58.4-47.4) AP margin over UDP for HRNet-W32 respectively, and 10.5\% (61.5-51.0) AP gain for HRNet-W48. Compared with DARK, CAL attains 11.1\% (58.4-47.3) and 4.1\% (61.5-57.4) AP gain for HRNet-W32 and HRNet-W48.
(ii) For the resolution of 128$\times$96, CAL achieves an extra margin of 4.8\% (74.3-69.5) AP as compared to UDP
for HRNet-W32, and 3.1\% (75.0-71.9) AP improvement for HRNet-W48. 
Compared with DARK, CAL obtains 1.5\% (74.3-72.8) and 0.9\% (75.0-74.1) AP improvement for HRNet-W32 and HRNet-W48.
Whilst DARK is a strong competitor particularly at relatively higher resolutions (\eg, 128$\times$96), it is still significantly inferior when the input resolution becomes smaller. This suggests that DARK's post-processing is less effective in tackling low-resolution input, compared to
the proposed learning based offset modeling.

\begin{table}[ht]
    \centering
    \scalebox{0.8}{
        \begin{tabular}{ c | c | c | c | c |c | c | c | c}
            \hline
            Method & Backbone & Input size & $AP$ & $AP^{50}$ & $AP^{75}$ & $AP^{M}$ & $AP^{L}$ & $AR$ \\
            \hline \hline
            HRNet \cite{HRNet:0009XLW19} & \multirow{4}{*}{HRNet-W32} & \multirow{4}{*}{64 $\times$ 48 } & 29.7 & 75.7 & 13.1 & 29.3 & 30.7 & 37.3 \\
            UDP \cite{UDP:huang2019devil} &         &         & 47.4 & 80.5 & 50.6 & 47.7 & 47.7 & 53.8 \\
            DARK \cite{DARK:Zhang2020} &   &   & 47.3 & 86.2 & 49.5 & 47.0 & 48.3 & 54.5 \\
           \bf CAL (Ours) &                            &                                   & \bf{58.4} & \bf{86.6} & \bf{65.1} & \bf{57.3} & \bf{60.5} & \bf{64.8} \\
            \hline
            HRNet& \multirow{4}{*}{HRNet-W48} & \multirow{4}{*}{64 $\times$ 48 } & 32.4 & 78.3 & 16.2 & 31.5 & 34.0 & 39.3 \\
            UDP  &                            &                                   & 51.0 & 82.6 & 55.2 & 51.4 & 51.0 & 57.3 \\
            DARK &                            &                                   & 57.4 & 87.1 & 64.8 & 56.5 & 58.9 & 62.0 \\ 
            \bf CAL (Ours) &                            &                                   & \bf{61.5} & \bf{88.1} & \bf{68.7} & \bf{60.7} & \bf{63.5} & \bf{66.3} \\  
            \hline
            HRNet& \multirow{4}{*}{HRNet-W32} & \multirow{4}{*}{128 $\times$ 96 } & 61.5 & 90.4 & 71.8 & 59.2 & 65.4 & 66.0 \\
            UDP  &                            &                                   & 69.5 & 91.3 & 77.7 & 67.9 & 72.2 & 73.7 \\
            DARK &                            &                                   & 72.8 & 92.5 & 80.5 & 70.7 & 76.5 & 75.9 \\
            \bf CAL (Ours) &                            &                                   & \bf{74.3} & \bf{92.6} & \bf{81.5} & \bf{72.1} & \bf{77.5} & \bf{77.2} \\
            \hline
            HRNet& \multirow{4}{*}{HRNet-W48} & \multirow{4}{*}{128 $\times$ 96 } & 62.9 & 91.4 & 73.8 & 60.2 & 66.7 & 67.2 \\
            UDP  &                            &                                   & 71.9 & 91.3 & 79.9 & 70.2 & 75.3 & 75.8 \\
            DARK &                            &                                   & 74.1 &\bf 92.6 & 81.5 & 71.9 & 77.8 & 77.0 \\
            \bf CAL (Ours) &                            &                                   & \bf{75.0} & \bf{92.6} & \bf{82.5} & \bf{72.6} & \bf{78.6} & \bf{77.8} \\
            \hline
        \end{tabular}
    }%
    \caption{Effect of low-resolution input on the COCO valid set.}
    \label{tab:resolution}
\end{table}

\subsection{High-resolution input setting}
To evaluate the effect of high-resolution input on the performance of human pose estimation,
we compared our method with the state-of-the-art methods in the conventional high-resolution setting.
Table \ref{tab:STOA_COCO_val} shows that CAL can still achieve performance superiority compared with the best competitor, UDP based HRNet-W48, although the margin is not as large as in the more challenging low-resolution setting.

\begin{table*}[htb]
	\setlength{\tabcolsep}{0.1cm}
	\begin{center}
		\resizebox{\columnwidth}{!}{%
		\begin{tabular}{ l | c | c | c | c | c | c | c | c | c | c  }
			\hline
			Method & Backbone & Input size & \#Params & GFLOPs & $AP$ & $AP^{50}$ & $AP^{75}$ & $AP^{M}$ & $AP^{L}$ & $AR$ \\
			\hline \hline
			IPR \cite{integral:SunXWLW18} & ResNet-101 & $256\times256$ & 45.1M & 11.0
			& 67.8 & 88.2 & 74.8 & 63.9 & 74.0 & -
			\\
			RMPE \cite{RMPE:Fang17} & PyraNet & $320\times256$ & 28.1M & 26.7
			& 72.3 & 89.2 & 79.1 & 68.0 & 78.6 & -
			\\
			CPN \cite{CPN:ChenWPZYS18} & ResNet-Inception & $384\times288$ & - & -
			& 72.1 & 91.4 & 80.0 & 68.7 & 77.2 & 78.5
			\\
			CPN (ensemble) \cite{CPN:ChenWPZYS18} & ResNet-Inception & $384\times288$ & - & -
			& 73.0 & 91.7 & 80.9 & 69.5 & 78.1 & 79.0 
			\\
			SimpleBaseline\cite{simplebaseline:XiaoWW18} & ResNet-152 & $384\times288$ & 68.6M & 35.6
			& 73.8 & 91.7 & 81.2 & 70.3 & 80.0 & 79.1
			\\ 
			HRNet & HRNet-W48 & $384\times288$ & 63.8M & 32.9
			& 77.1 & 91.8 & 83.8 & 73.5 & 83.5 & 81.8
			\\
			UDP & HRNet-W48 & $384 \times 288$ & 63.6M & 33.0
			& 77.8 & 92.0 & 84.3 & 74.2 & 84.5 & 82.5
			\\
			\hline
			\bf CAL (Ours) & HRNet-W48 & $384 \times 288$ & 110.2M & 47.2
			& \bf{78.2} & \bf{92.1} & \bf{84.6} & \bf{74.5} & \bf{85.0} & \bf{82.8}
			\\
			\hline
		\end{tabular}
		}
	\end{center}
	\caption{
		Results in conventional high-resolution input setting on the COCO valid set. 
	}
	\label{tab:STOA_COCO_val}
\end{table*}

\section{Conclusion}
\label{sec:conclusion}
In this paper, we propose a novel {\em Confidence-Aware Learning} (CAL) method for low-resolution human pose estimation, a largely ignored problem in the literature. 
This is established on our solid investigation of existing methods
for low-resolution challenges which reveals that offset learning is an effective approach.
In particular, CAL addresses two fundamental limitations 
of previous offset-based methods by introducing 
weighing masks to the learning process of both heatmap
and offset. This makes the model training and testing 
more consistent while coupling the two learning targets
more closely.
Furthermore, a two-stage training strategy for our method is developed to boost the performance.
Comprehensive experiments on the popular COCO benchmark show the superiority of our CAL method over existing state-of-the-art methods
for low-resolution pose estimation.
Finally, ablation studies are carried out to evaluate the 
effectiveness of individual components of our method.

In the further work, to make our method more scalable and deployable to a wider variety of computing environments, we aim to further reduce the computational complexity of our CAL method by leveraging additional techniques such as lightweight networks, sparse convolution, network pruning, quantization, and neural architecture search.

\biboptions{numbers,sort&compress}
\bibliography{mybibfile}

\end{document}